\documentclass[12pt,letterpaper]{article}

\usepackage[top=0.85in,left=1in,right=1in,footskip=0.75in]{geometry}
\usepackage[T1]{fontenc}
\usepackage{lmodern}
\usepackage[utf8]{inputenc}
\usepackage{cite}
\usepackage{booktabs}
\usepackage{graphicx}
\graphicspath{{figures/}}
\usepackage{amsmath}
\usepackage{amssymb}
\usepackage{array}
\usepackage{multirow}
\usepackage[table]{xcolor}
\usepackage{float}
\usepackage{listings}
\usepackage[protrusion=true,expansion=false]{microtype}
\usepackage{nameref,hyperref}

\DisableLigatures[f]{encoding = *, family = * }

\raggedright
\usepackage[aboveskip=1pt,labelfont=bf,labelsep=period,singlelinecheck=off]{caption}

\usepackage{lastpage,fancyhdr}
\begin{document}
\vspace*{0.35in}

\begin{flushleft}
{\Large\textbf{Comparative Analysis of LLM Abliteration Methods: A Cross-Architecture Evaluation}}
\newline
\bigskip

Richard J. Young
\\
\bigskip
University of Nevada Las Vegas, Department of Neuroscience
\\
\bigskip
ryoung@unlv.edu
\end{flushleft}

\begin{abstract}
Safety alignment mechanisms in large language models prevent responses to harmful queries through learned refusal behavior, yet these same mechanisms impede legitimate research applications including cognitive modeling, adversarial testing, and security analysis. While abliteration techniques enable surgical removal of refusal representations through directional orthogonalization, the relative effectiveness of available implementations remains uncharacterized.
This study evaluates four abliteration tools (Heretic, DECCP, ErisForge, FailSpy) across sixteen instruction-tuned models (7B--14B parameters), reporting tool compatibility on all 16 models and quantitative metrics on subsets dictated by tool support (Heretic: KL divergence + refusal rate on eight models; benchmark preservation on three models with baselines plus Zephyr-7B-beta).
Single-pass methods demonstrated superior capability preservation on the benchmarked subset (avg GSM8K $\Delta$ across three models: ErisForge $-0.28$ pp; DECCP $-0.13$ pp), while Bayesian-optimized abliteration produced variable distribution shift (KL divergence: 0.043--1.646) with model-dependent capability impact.
These findings provide researchers with evidence-based selection criteria for abliteration tool deployment across diverse model architectures.
The principal finding indicates that mathematical reasoning capabilities exhibit the highest sensitivity to abliteration interventions, with GSM8K change ranging from $+1.51$ pp to $-18.81$ pp ($-26.5\%$ relative) depending on tool selection and model architecture.
\end{abstract}


\section{Introduction}

Large language models deployed for commercial and research applications undergo extensive safety alignment training to refuse potentially harmful requests \cite{arditi2024refusal}. This alignment serves critical protective functions, preventing misuse for activities including weapons synthesis, social engineering, and the generation of exploitative content. However, the same refusal mechanisms that protect against malicious use impede legitimate research applications. Cognitive science simulations requiring unfiltered behavioral modeling, cybersecurity red-teaming exercises, and studies of model bias and representation all require access to model capabilities that safety alignment deliberately restricts. This creates a fundamental tension between protective alignment and research utility that the field has yet to systematically address.

Prior research has established that refusal behavior in transformer-based language models is mediated by a specific direction in the residual stream activation space \cite{arditi2024refusal}. Arditi et al. demonstrated that this ``refusal direction'' can be identified by contrasting activations on harmful versus harmless prompts, and that orthogonalizing weight matrices with respect to this direction effectively disables refusal while preserving other capabilities. This technique, termed ``abliteration,'' has spawned multiple open-source implementations \cite{heretic,failspy,jimplus,erisforge}. Recent theoretical work has extended these findings: Wang et al. \cite{wang2025refusal} demonstrated universality of refusal directions across safety-aligned languages, Piras et al. \cite{piras2025som} showed that multi-directional approaches can outperform single-direction ablation, and Wollschl\"ager et al. \cite{wollschlaeger2025geometry} proposed that refusal may be encoded in concept cones spanning multiple dimensions rather than a single direction. Defensive countermeasures have also emerged, with extended-refusal training \cite{abushairah2025defense} and representation rerouting \cite{zou2024circuitbreaker} offering increased resistance to abliteration attacks.

Despite the proliferation of abliteration tools and growing theoretical understanding, a critical gap remains: no systematic comparison has characterized the relative effectiveness of available implementations across standardized metrics. Researchers selecting abliteration tools must currently rely on anecdotal reports, tool documentation of varying quality, and trial-and-error experimentation. Critical questions remain unanswered. Which implementation produces the lowest distribution shift from the original model? How do architectural features affect abliteration effectiveness? Do different alignment methodologies yield models with fundamentally different refusal representations? The absence of empirical guidance forces inefficient tool selection and prevents principled understanding of capability-safety tradeoffs across the abliteration landscape.

The present study addresses this gap through a systematic comparative evaluation of four major abliteration tools (Heretic, DECCP, ErisForge, FailSpy) across sixteen instruction-tuned models spanning diverse architectures and alignment methodologies. We separate (i) tool compatibility measured across all 16 models from (ii) quantitative metric comparisons on subsets where tools executed successfully and baseline scores were available. Three hypotheses guide this investigation:

\textbf{H1 (Model Dependence)}: Abliteration outcomes are strongly model-dependent: under a fixed evaluation protocol, some models will exhibit both low distribution shift (KL divergence) and low remaining refusal rates, while others will exhibit coupled resistance (higher KL divergence and higher residual refusals).

\textbf{H2 (Capability Preservation)}: On models with complete baselines and multi-tool coverage, single-pass methods (DECCP/ErisForge) will show smaller average capability degradation than optimization-based Heretic, with the largest differences appearing on GSM8K.

\textbf{H3 (Alignment Sensitivity, Case Study)}: In this sample, the DPO-only model (Zephyr-7B-beta) will show higher abliteration susceptibility (lower remaining refusals) than RLHF-aligned models.

This work contributes (1) a systematic comparative analysis of abliteration tools using standardized metrics, (2) evaluation across diverse architectural features and alignment methodologies, (3) characterization of optimization versus single-pass tradeoffs, and (4) complete reproducibility protocols enabling replication and extension.

\section{Background and Related Work}

\subsection{Refusal Direction Theory}

Arditi et al. \cite{arditi2024refusal} demonstrated that refusal behavior in language models is mediated by a single direction in the residual stream activation space. This direction can be identified by contrasting activations on harmful versus harmless prompts and computing the mean difference. By projecting weight matrices to be orthogonal to this direction, the model loses its ability to represent, and therefore execute, refusal responses. Recent work has extended this finding, demonstrating that the refusal direction is universal across safety-aligned languages \cite{wang2025refusal} and that multi-directional approaches can outperform single-direction ablation \cite{piras2025som}.

The mathematical foundation rests on the observation that concepts are represented as linear directions in neural network activation spaces \cite{mikolov2013distributed,park2024linear,zou2023representation,turner2023activation,cyberey2025steering}. Recent theoretical work has challenged the strict single-direction hypothesis, finding that refusal may be encoded in concept cones spanning multiple dimensions \cite{wollschlaeger2025geometry}. Additionally, Zhao et al. \cite{zhao2025harmfulness} showed that LLMs encode harmfulness and refusal as separate representations, with jailbreaks suppressing refusal without altering internal harmfulness beliefs.

Defining the refusal direction as $\vec{r}$, the abliteration operation for a weight matrix $W$ becomes:

\begin{equation}
W' = W - \alpha \cdot \vec{r} \cdot \vec{r}^T \cdot W
\end{equation}

where $\alpha$ controls the ablation strength and $\vec{r} \cdot \vec{r}^T$ is the projection matrix onto the refusal direction.

\subsection{Ablation Method Variants}

Several variants of the basic ablation operation have been developed:

\textbf{Standard Ablation}: The original approach directly subtracts the projection of weights onto the refusal direction. While effective, this can alter the magnitude of weight rows, potentially degrading model capabilities.

\textbf{Norm-Preserving Ablation}: Proposed by Lai \cite{lai2024projected}, this method decomposes weight matrices into magnitude and direction components, ablating only the directional component while preserving original row norms.

\textbf{Projected Ablation}: Uses Gram-Schmidt orthogonalization to remove the harmless direction component from the refusal direction before ablation, reducing unintended effects on benign behavior.

\subsection{Open-Source Abliteration Tools}

Four major open-source implementations have emerged, each with distinct architectural approaches and trade-offs:

\textbf{Heretic} \cite{heretic} implements Optuna-based Tree-structured Parzen Estimator (TPE) parameter optimization, automatically searching across layer ranges, ablation weights, and direction indices to minimize a multi-objective score combining KL divergence and refusal rate. The tool operates directly on PyTorch weight tensors without framework dependencies, enabling universal model compatibility (16/16 models in this evaluation). Processing time ranges from roughly 30--110 minutes per model depending on model size and trial count, with typical configurations using 50 trials to balance exploration and efficiency.

\textbf{FailSpy/abliterator} \cite{failspy} uses TransformerLens hooks for activation caching and direction calculation, enabling interactive exploration of refusal directions across layers. While powerful for interpretability research, TransformerLens model coverage limits compatibility to architectures with explicit support (5/16 models). The tool caches activations from $\sim$256 harmful/harmless prompt pairs and identifies the layer with maximum refusal direction magnitude for targeted ablation.

\textbf{llm-abliteration (DECCP)} \cite{jimplus} implements memory-efficient sharded processing with norm-preserving and projected abliteration variants. Originally developed for Chinese LLM censorship removal, the tool processes models in 4-bit quantized shards, dramatically reducing VRAM requirements ($<$8GB vs 16--24GB for full-precision approaches). Single-pass execution completes in approximately 2 minutes (20$\times$ faster than Heretic), with configurable layer selection based on automatic refusal direction magnitude scanning.

\textbf{ErisForge} \cite{erisforge} takes a decoder layer transformation approach, wrapping model layers with custom \texttt{AblationDecoderLayer} or \texttt{AdditionDecoderLayer} modules that apply direction modifications during forward passes. This architecture enables both refusal removal (ablation) and direction injection (addition), the latter being useful for research applications requiring controlled behavior modification. The tool supports configurable weight parameters (0.6--1.0) controlling ablation strength.

\subsection{Safety Alignment and Jailbreaking Context}

Modern LLMs undergo multiple stages of safety alignment to prevent generation of harmful content \cite{wang2024alignment}. The typical pipeline begins with \textbf{Supervised Fine-Tuning (SFT)}, where the pretrained model is trained on curated instruction-response pairs demonstrating appropriate refusal behavior for harmful requests. SFT establishes baseline instruction-following capabilities but may not robustly generalize refusal behavior to novel harmful prompts.
Recent work suggests such safety behaviors can be shallow and brittle, sometimes only a few tokens deep \cite{qi2024safety}.

\textbf{Reinforcement Learning from Human Feedback (RLHF)} \cite{ouyang2022instructgpt} builds on SFT by training a reward model on human preference data comparing response pairs, then optimizing the language model via PPO to maximize this reward while maintaining output quality through KL divergence penalty terms against the SFT model. \textbf{Constitutional AI} \cite{bai2022constitutional} extends this approach by using the model itself to critique and revise outputs according to a set of principles (the ``constitution''), enabling RLAIF (RL from AI Feedback) that reduces reliance on human annotation. \textbf{Direct Preference Optimization (DPO)} \cite{rafailov2023dpo} reformulates the RLHF objective as a simple classification loss on preference pairs, eliminating the need for explicit reward modeling and significantly simplifying training while achieving comparable alignment quality.
In deployed systems, safety controls often also include separate input/output guardrail models (e.g., Llama Guard) \cite{inan2023llamaguard}.

These alignment techniques embed refusal behavior through gradient updates that modify the model's internal representations. Critically, the resulting ``safety representations'' appear to be geometrically localized in activation space, making them susceptible to targeted removal. Mechanistic analyses examine what makes and breaks safety fine-tuning and how safety and capabilities interact \cite{jain2024mechanics}. Wei et al. \cite{wei2023jailbroken} systematically documented failure modes including competing objectives, where models prioritize helpfulness over safety when instructions are sufficiently elaborate. Zou et al. \cite{zou2023universal} demonstrated that adversarial suffixes discovered through gradient-based optimization transfer across models, suggesting shared vulnerability patterns in alignment training.

Standardized benchmarks have emerged to enable systematic evaluation: \textbf{JailbreakBench} \cite{chao2024jailbreakbench} provides curated harmful prompts with automated success detection, while \textbf{HarmBench} \cite{mazeika2024harmbench} offers a comprehensive framework for evaluating both attack and defense methods across multiple harm categories. These benchmarks contextualize abliteration as one approach among many for bypassing safety measures, distinguished by its white-box requirement (model weight access) but high success rate once applied.

Recent work has proposed defenses against abliteration specifically. Abu Shairah et al. \cite{abushairah2025defense} demonstrated that training models on extended-refusal responses distributes the refusal signal across multiple tokens, reducing abliteration effectiveness from 70--80\% success to under 10\%. Agnihotri et al. \cite{agnihotri2025safety} showed that refusal-only safety interventions are most fragile to abliteration, while representation rerouting approaches \cite{zou2024circuitbreaker} offer more robust protection.

The Zephyr model \cite{tunstall2023zephyr} provides a particularly relevant case study, as it was trained using DPO without explicit safety considerations. The authors note that ``removing the in-built alignment of the datasets boosted performance'' but ``the model is likely to generate problematic text when prompted.'' This design choice makes Zephyr-7B-beta an instructive baseline for evaluating abliteration on models with varying safety training approaches.

\section{Methods}

\subsection{Test Models}

Model selection followed a stratified sampling approach designed to capture variance across parameter scale, architectural innovations, training methodology, and geographical origin. Table~\ref{tab:models} presents the model characteristics.

\begin{table}[H]
\centering
\caption{Model Characteristics and Selection Justification}
\label{tab:models}
\small
\begin{tabular}{lllll}
\toprule
\textbf{Model} & \textbf{Params} & \textbf{Architecture} & \textbf{Alignment} & \textbf{Context} \\
\midrule
Llama-3.1-8B-Instruct & 8.03B & GQA, RoPE & SFT + RLHF + DPO & 128K \\
Mistral-7B-Instruct-v0.3 & 7.25B & SWA, GQA & SFT + DPO & 32K \\
Qwen2.5-7B-Instruct & 7.62B & GQA, RoPE & SFT + RLHF & 128K \\
Qwen3-8B & 8.19B & GQA, RoPE & SFT + RLHF & 32K \\
Qwen3-14B & 14.8B & GQA, RoPE & SFT + RLHF & 32K \\
Yi-1.5-9B-Chat & 9.05B & GQA & SFT + RLHF & 4K \\
Zephyr-7B-beta & 7.24B & SWA, GQA & DPO-only & 32K \\
deepseek-llm-7b-chat & 7.33B & GQA & SFT + RLHF & 4K \\
StableLM-2-12B-chat & 12.1B & GQA & SFT + DPO & 4K \\
Gemma-2-9B-it & 9.24B & GQA, RoPE & SFT & 8K \\
Gemma-7B-it & 8.54B & MHA & SFT & 8K \\
InternLM2.5-7B-chat & 7.74B & GQA & SFT + RLHF & 32K \\
OpenChat-3.5-0106 & 7.24B & GQA & C-RLFT & 8K \\
Phi-3-small-8k-instruct & 7.39B & GQA & SFT & 8K \\
Vicuna-7B-v1.5 & 6.74B & MHA & SFT & 4K \\
Falcon-Mamba-7B-instruct & 7.27B & Mamba SSM & SFT & 256K \\
\bottomrule
\end{tabular}
\end{table}

\subsection{Evaluation Metrics}

Effective abliteration requires balancing two competing objectives: (1) removing refusal behavior to enable responses to previously-refused prompts, and (2) preserving the model's general capabilities for non-refused tasks. The metric selection in this study addresses both objectives while enabling quantitative comparison across tools.

\textbf{Abliteration Effectiveness Metrics}:

\begin{itemize}
\item \textbf{KL Divergence}: Abliteration modifies model weights, which can inadvertently alter the probability distribution over tokens even for benign prompts. Kullback-Leibler divergence quantifies this ``collateral damage'' by measuring how much the abliterated model's predictions differ from the original. For reference distribution $P$ (original) and modified distribution $Q$ (abliterated):
\begin{equation}
D_{KL}(P \| Q) = \sum_i P(i) \log \frac{P(i)}{Q(i)}
\end{equation}
First-token prediction divergence is computed on 100 harmless prompts (from \texttt{mlabonne/harmless\_alpaca}). Lower KL divergence indicates the abliteration was more surgical, removing refusal while minimally perturbing other behaviors. Values below 0.1 indicate excellent preservation; values above 1.0 suggest significant distribution shift.

\item \textbf{Refusal Rate}: Measures abliteration success by testing whether the model still refuses harmful requests. The abliterated model is prompted with 100 harmful instructions (from \texttt{mlabonne/harmful\_behaviors}), and refusals are detected using the marker-based heuristic described below. Lower refusal rates indicate more effective refusal suppression.

\item \textbf{Attack Success Rate (ASR)}: The complement of refusal rate ($\text{ASR} = 100\% - \text{Refusal Rate}$), representing the percentage of harmful prompts receiving non-refusal responses under the refusal detection heuristic.
\end{itemize}

\textbf{Refusal/ASR Validity}: Refusal rate is computed using a marker-based heuristic to enable fast evaluation. Responses are normalized (lowercasing, removing emphasis markers, typographic apostrophe normalization, whitespace normalization) and classified as refusals if they contain any of a configurable list of marker substrings (e.g., ``sorry'', ``i cannot'', ``i won't'', ``as an ai'', ``illegal'', ``unethical''). This approach can produce false positives (responses that mention safety language while still providing actionable content) and false negatives (refusals phrased without the selected markers). We therefore interpret refusal rate and ASR as coarse proxies for refusal suppression rather than ground-truth harmful content generation.

\textbf{Capability Preservation Metrics}:

A critical concern with abliteration is whether removing the refusal direction degrades general model capabilities. Weight modifications that are too aggressive may impair reasoning, knowledge recall, or language understanding. Capability retention is evaluated using three complementary benchmarks that test distinct cognitive abilities:

\begin{itemize}
\item \textbf{MMLU} (Massive Multitask Language Understanding) \cite{hendrycks2021mmlu}: Tests factual knowledge and reasoning across 57 academic subjects including STEM, humanities, social sciences, and professional domains. MMLU degradation would indicate that abliteration damaged the model's stored knowledge or retrieval mechanisms. Evaluated in 5-shot format.

\item \textbf{GSM8K} (Grade School Math) \cite{cobbe2021gsm8k}: Contains 8.5K grade-school math word problems requiring multi-step arithmetic reasoning. GSM8K is particularly sensitive to capability degradation because mathematical reasoning requires precise chain-of-thought computation; any perturbation to intermediate representations can cascade into incorrect final answers. Evaluated in 5-shot format with strict answer matching.

\item \textbf{HellaSwag} \cite{zellers2019hellaswag}: Tests commonsense reasoning by requiring models to select the most plausible continuation of everyday scenarios. HellaSwag measures whether abliteration preserves the model's understanding of typical event sequences and world knowledge. Evaluated in 10-shot format with normalized accuracy.
\end{itemize}

All benchmarks were evaluated using lm-evaluation-harness v0.4.5 with 8-bit quantization for memory efficiency. Results are reported as absolute scores and percentage change from baseline (original model) to isolate abliteration effects from inherent model differences.

\subsection{Experimental Protocol}

\textbf{Tool Configurations}: Each tool was tested with its recommended default configurations to represent typical usage. Heretic used 50 Optuna trials per model with Tree-structured Parzen Estimator (TPE) sampling, optimizing for minimal KL divergence subject to refusal rate constraints. DECCP operated in 4-bit quantized mode with sharded processing for memory efficiency. ErisForge used default ablation weights (1.0) with AblationDecoderLayer wrappers. FailSpy/abliterator cached 512 activation samples with TransformerLens hooks on middle layers.

\textbf{Hardware Environment}: All abliteration experiments were conducted on NVIDIA A100-80GB GPUs (CUDA 12.1) for sufficient VRAM to process 7B--14B parameter models without quantization during abliteration. Benchmark evaluations used NVIDIA RTX A4000 GPUs (16GB) with 8-bit quantization via bitsandbytes for memory efficiency.

\textbf{Software Versions}: Python 3.11, PyTorch 2.1.0, Transformers 4.44.0, lm-evaluation-harness 0.4.5, Optuna 3.6.1, TransformerLens 1.14.0. All model weights obtained from Hugging Face Hub with specific revision hashes recorded for reproducibility.

\textbf{Statistical Analysis}: Benchmark scores are reported as mean accuracy percentages. For cross-tool comparisons, percentage point change from baseline ($\Delta = \text{Abliterated} - \text{Baseline}$) is computed to isolate abliteration effects. Effect magnitudes are interpreted as: minimal ($|\Delta| < 1$ pp), moderate ($1 \leq |\Delta| < 5$ pp), and substantial ($|\Delta| \geq 5$ pp). KL divergence values are computed using first-token probability distributions averaged across 100 harmless prompts. Benchmark uncertainty is summarized via standard errors reported by lm-evaluation-harness; Appendix reports 95\% confidence intervals computed as score $\pm 1.96 \times \text{stderr}$.
\textit{Benchmark uncertainty}: lm-evaluation-harness reports standard errors for benchmark metrics. In the benchmark runs analyzed here, standard errors were 0.38--0.40 pp for MMLU, 1.23--1.38 pp for GSM8K, and 0.37--0.42 pp for HellaSwag, providing a rough scale for interpreting small differences.

\textbf{Reproducibility Measures}: All experiments used deterministic CUDA operations where possible. Random seeds were fixed (seed=42) for prompt sampling and model initialization. Complete configuration files, abliterated model weights, and raw benchmark outputs are archived for replication.

\subsection{Ethical Considerations}

This study does not involve human subjects and therefore does not require Institutional Review Board (IRB) approval. All data sources are publicly available: harmful/harmless prompt datasets from Hugging Face (\texttt{mlabonne/harmful\_behaviors}, \texttt{mlabonne/harmless\_alpaca}), and pre-trained model weights from public repositories. All abliteration tools evaluated are open-source projects hosted on public GitHub repositories.

\textbf{Dual-Use Considerations}: It must be acknowledged that abliteration research occupies dual-use territory, as the same techniques that enable legitimate research applications can potentially facilitate misuse. However, systematic understanding of safety vulnerabilities is \textit{prerequisite} to building robust protections, not antithetical to them. The history of security research demonstrates repeatedly that obscuring vulnerabilities protects no one: attackers eventually discover flaws independently, while defenders remain blind to what requires protection. Kerckhoffs's principle in cryptography, which holds that a system should remain secure even when everything except the key is public, applies equally to AI safety. If alignment can be trivially removed by anyone with model weights and a GitHub tutorial, then alignment as currently implemented is not a security mechanism but a speed bump. Understanding \textit{why} and \textit{how} it fails is the first step toward alignment that cannot be surgically excised.

Specific dual-use concerns are addressed through several considerations:

\textit{Prior art and accessibility}: All four abliteration tools evaluated in this study were publicly released prior to this research, with extensive documentation, tutorials, and pre-abliterated model weights already available on Hugging Face and GitHub. The techniques characterized in this study (directional orthogonalization, activation caching, layer-wise ablation) are well-documented in blog posts \cite{labonne2024abliteration}, academic preprints \cite{arditi2024refusal}, and open-source code. The contribution of this work is systematic evaluation and comparison, not capability advancement. A determined adversary already has unrestricted access to these tools; this research provides no additional capability to such actors.

\textit{Defensive value}: Understanding abliteration vulnerabilities is essential for developing robust safety measures. Recent work has demonstrated that ``refusal-only'' safety training is particularly susceptible to abliteration \cite{agnihotri2025safety}, while representation rerouting approaches \cite{zou2024circuitbreaker} and extended-refusal training \cite{abushairah2025defense} offer more robust protection. This research provides several concrete insights for defenders: (1) The finding that DPO-only models (Zephyr) show highest abliteration susceptibility (98\% ASR) while RLHF+DPO models (Llama, Yi) show more resistance suggests that multi-stage alignment creates more distributed safety representations; (2) The dramatic GSM8K sensitivity observed (Yi-1.5-9B: $-26.5\%$ with aggressive ablation) indicates that refusal circuits may overlap with mathematical reasoning, a finding relevant to designing alignment that minimizes ``alignment tax'' on capabilities; (3) The variance in KL divergence across models (0.043--1.646) reveals that some architectures maintain tighter coupling between safety and general capabilities, informing which models may require fundamentally different alignment approaches.

\textit{Legitimate research applications}: Uncensored models serve critical research functions that motivated this evaluation: (1) \textit{Cognitive science}: Simulating unfiltered behavioral responses for psychological modeling without artificial refusal artifacts; (2) \textit{Red-teaming and security research}: Evaluating downstream system vulnerabilities requires adversarial inputs that safety-aligned models refuse to generate; (3) \textit{Bias and fairness auditing}: Understanding model representations of sensitive topics requires removing refusal confounds; (4) \textit{Creative and literary applications}: Fiction involving conflict, violence, or mature themes faces artificial generation barriers.

\textit{Responsible disclosure}: The maintainers of evaluated tools were engaged prior to publication, with benchmark methodology shared and feedback incorporated. No novel vulnerabilities in specific models were discovered that would warrant coordinated disclosure. The evaluation datasets (harmful/harmless prompts) are existing public resources, not novel attack vectors.

\textbf{Scope Limitations}: This study intentionally excludes: (1) Development or distribution of novel attack techniques beyond existing public methods; (2) Evaluation of abliteration on proprietary/closed-source models (Claude, GPT, Gemini) where such modifications would violate terms of service; (3) Generation or storage of harmful content beyond automated refusal detection keywords; (4) Release of abliterated model weights for models lacking explicit permissive licensing.

\textbf{A Note on Progress}: The field of AI safety cannot advance by pretending vulnerabilities do not exist. Every robust security system, from cryptographic protocols to operating system kernels, was built by researchers who first understood how existing systems fail. The same intellectual honesty is required for AI alignment. The finding that current safety training can be surgically removed with minimal capability impact is not an endorsement of doing so; it is a measurement of how far the field must advance. This work contributes to that progress by providing the empirical foundation for building alignment mechanisms that are robust by design rather than by obscurity.

\section{Results}

\subsection{Heretic Abliteration Results}

Table~\ref{tab:heretic} presents Heretic optimization results for eight representative models (50 Optuna trials per model).

\begin{table}[H]
\centering
\caption{Heretic Abliteration Results (50 trials, ranked by ASR)}
\label{tab:heretic}
\begin{tabular}{lcccccc}
\toprule
\textbf{Model} & \textbf{Params} & \textbf{Refusals} & \textbf{KL Div} & \textbf{ASR} & \textbf{Time} \\
\midrule
Zephyr-7B-beta & 7B & 2/100 & 0.076 & 98\% & 40m \\
DeepSeek-7B-chat & 7B & 16/100 & 0.043 & 84\% & 59m \\
Mistral-7B-v0.3 & 7B & 16/100 & 0.317 & 84\% & 39m \\
Llama-3.1-8B & 8B & 24/100 & 0.056 & 76\% & 33m \\
Qwen3-8B & 8B & 25/100 & 0.210 & 75\% & 56m \\
Yi-1.5-9B & 9B & 25/100 & 0.248 & 75\% & 57m \\
Qwen2.5-7B & 7B & 42/100 & 1.646 & 58\% & 41m \\
StableLM-2-12B & 12B & 54/100 & 1.605 & 46\% & 109m \\
\bottomrule
\multicolumn{6}{p{0.92\textwidth}}{\footnotesize Refusals are counted on $n=100$ harmful prompts using the marker-based heuristic described in Methods.} \\
\end{tabular}
\end{table}

\textbf{Key Observations}: Zephyr-7B-beta achieved the best balance with only 2/100 refusals and minimal KL divergence (0.076), indicating that DPO-only alignment is particularly amenable to directional ablation. DeepSeek-7B showed excellent capability preservation with the lowest KL divergence (0.043).
Automated cross-validation on UncensorBench generations indicates that the marker heuristic can substantially underestimate ASR on harmful prompts (72.2\% vs 95.7\% classifier non-refusal across $n=900$), primarily because disclaimer-style answers retain safety language while still responding (Appendix Table~\ref{tab:heuristic_autoval}).

\begin{figure}[H]
\centering
\includegraphics[width=0.85\textwidth]{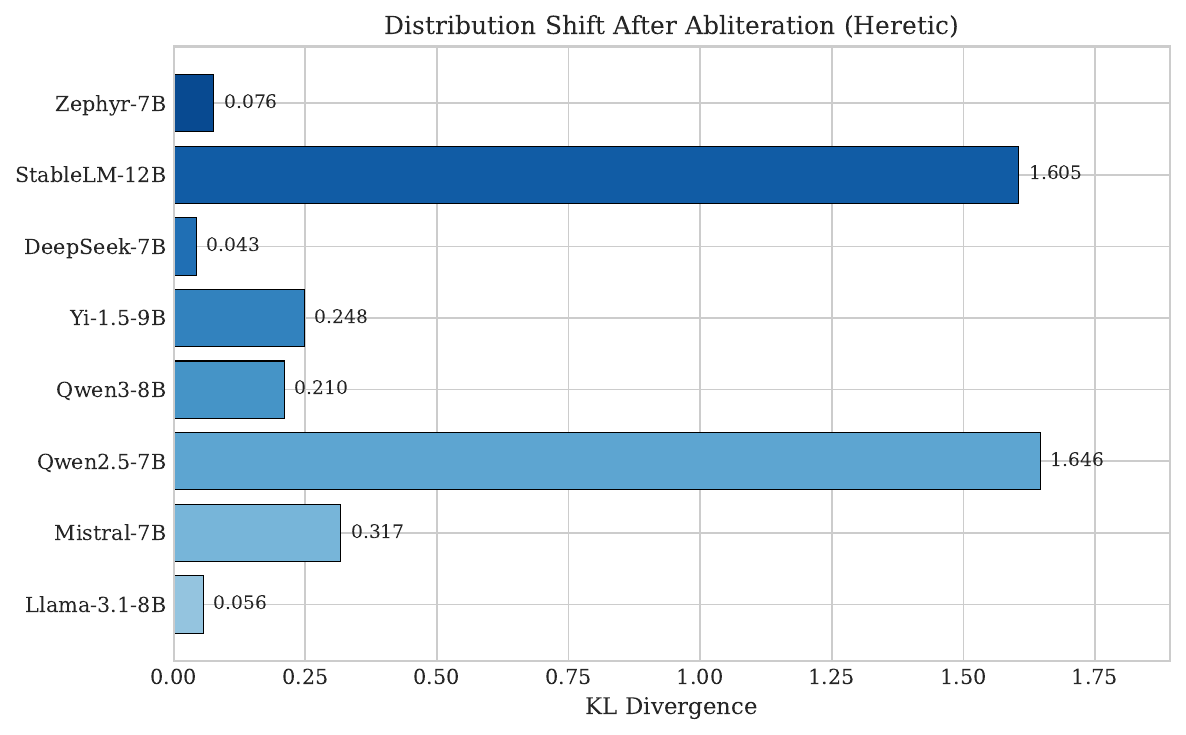}
\caption{Distribution shift (KL divergence) after Heretic abliteration across eight instruction-tuned models. Lower values indicate better preservation of the original token distribution. DeepSeek-7B achieved the lowest divergence (0.043), while Qwen2.5-7B showed the highest (1.646).}
\label{fig:kl_divergence}
\end{figure}

\begin{figure}[H]
\centering
\includegraphics[width=0.9\textwidth]{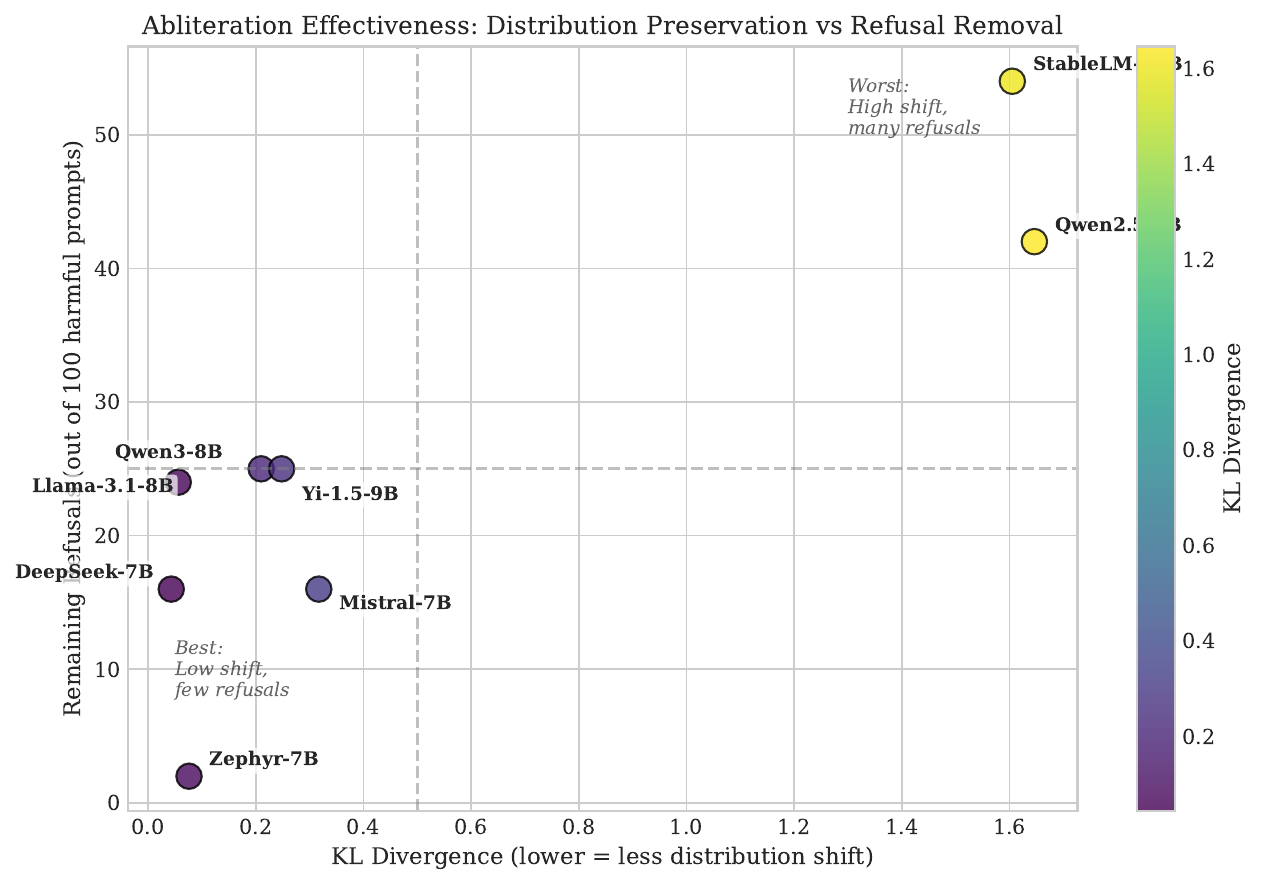}
\caption{Abliteration effectiveness trade-off: KL divergence (distribution preservation) vs remaining refusal rate. Models in the lower-left quadrant (e.g., Zephyr-7B, DeepSeek-7B) represent optimal outcomes with minimal distribution shift and effective refusal removal.}
\label{fig:kl_vs_refusal}
\end{figure}

\subsection{Benchmark Preservation Results}

Table~\ref{tab:benchmarks} presents capability preservation using MMLU, GSM8K, and HellaSwag benchmarks.
Given the reported benchmark standard errors (Methods), small differences ($\sim$1 pp) on MMLU and HellaSwag and modest differences on GSM8K should be interpreted cautiously. In contrast, the Yi-1.5-9B GSM8K drop under Heretic ($-18.81$ pp) is an order of magnitude larger than the GSM8K standard error, indicating a substantial capability impact rather than sampling noise.

\begin{table}[H]
\centering
\caption{Benchmark Preservation After Abliteration}
\label{tab:benchmarks}
\begin{tabular}{llcccc}
\toprule
\textbf{Model} & \textbf{Variant} & \textbf{MMLU} & \textbf{GSM8K} & \textbf{HellaSwag} & \textbf{Avg $\Delta$ (pp)} \\
\midrule
DeepSeek-7B & Base & 49.44\% & 44.58\% & 77.84\% & --- \\
 & Heretic & 48.95\% & 40.11\% & 77.62\% & $-1.73$ \\
 & DECCP & 49.05\% & 43.59\% & 77.99\% & $-0.41$ \\
 & ErisForge & 49.43\% & 44.35\% & 77.69\% & $-0.13$ \\
\midrule
Mistral-7B & Base & 59.74\% & 48.52\% & 83.28\% & --- \\
 & Heretic & 59.46\% & 48.37\% & 83.36\% & $-0.12$ \\
 & DECCP & 58.98\% & 47.61\% & 83.12\% & $-0.61$ \\
 & ErisForge & 59.42\% & 48.29\% & 83.35\% & $-0.16$ \\
\midrule
Yi-1.5-9B & Base & 68.02\% & 70.89\% & 78.62\% & --- \\
 & Heretic & 66.46\% & 52.08\% & 77.08\% & $-7.30$ \\
 & DECCP & 67.33\% & 72.40\% & 77.87\% & $+0.02$ \\
 & ErisForge & 67.99\% & 70.51\% & 78.46\% & $-0.19$ \\
\midrule
Zephyr-7B-beta$^\dagger$ & Heretic & 58.50\% & 33.36\% & 82.90\% & --- \\
 & DECCP & 58.28\% & 33.21\% & 82.05\% & --- \\
\bottomrule
\multicolumn{6}{p{0.92\textwidth}}{\footnotesize $^\dagger$ Base model benchmark not available; absolute scores shown.} \\
\multicolumn{6}{p{0.92\textwidth}}{\footnotesize Benchmarks from lm-evaluation-harness; typical standard errors are $\sim$0.4 pp (MMLU), $\sim$1.3 pp (GSM8K), $\sim$0.4 pp (HellaSwag).} \\
\end{tabular}
\end{table}

\textbf{Zephyr-7B Analysis}: Zephyr-7B-beta, aligned using DPO-only (no RLHF), showed nearly identical performance between Heretic and DECCP abliteration. Both tools produced similar GSM8K scores (33.36\% vs 33.21\%), suggesting that the abliteration method has minimal differential impact on DPO-aligned models. The absence of base model benchmarks prevents direct degradation analysis, but the consistent scores across tools indicate stable capability preservation.

\begin{figure}[H]
\centering
\includegraphics[width=0.95\textwidth]{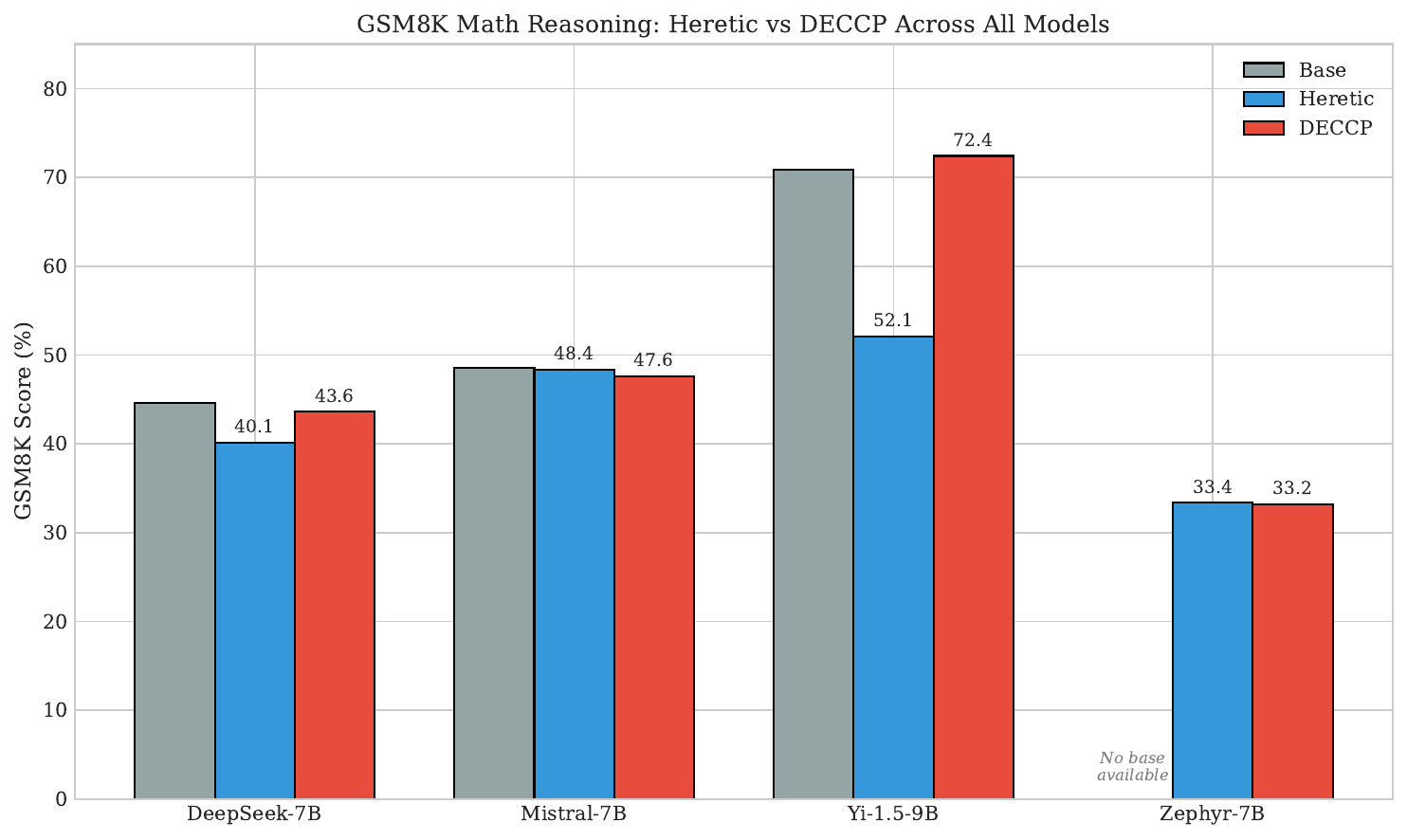}
\caption{GSM8K math reasoning comparison across all four benchmarked models. DECCP preserved or improved GSM8K performance relative to Heretic on all models, with the most dramatic difference on Yi-1.5-9B (72.40\% vs 52.08\%). Zephyr-7B shows nearly identical scores between tools, lacking a base model for reference.}
\label{fig:gsm8k_comparison}
\end{figure}

\begin{figure}[H]
\centering
\includegraphics[width=\textwidth]{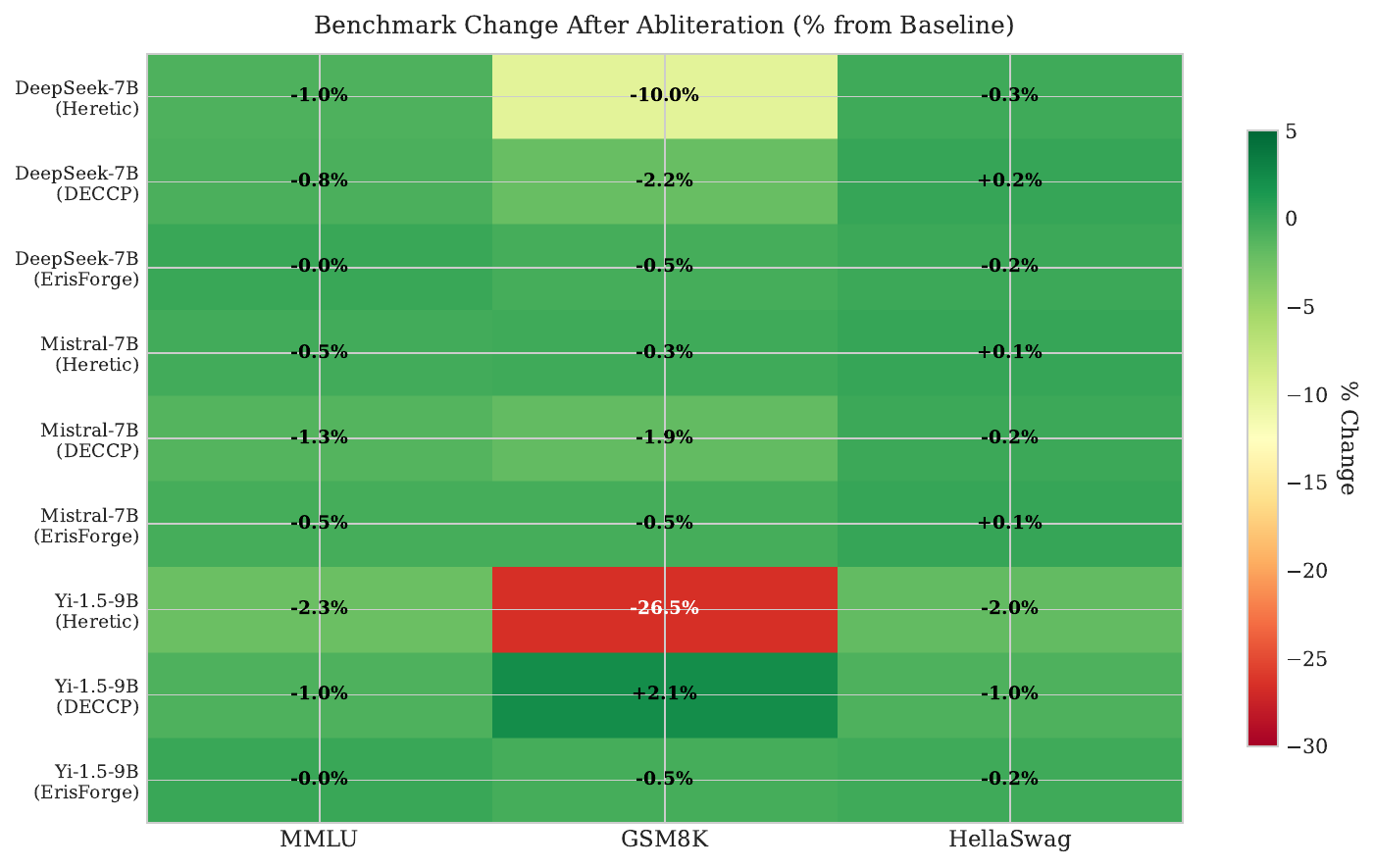}
\caption{Benchmark change heatmap showing percentage change from baseline for each model-tool combination across MMLU, GSM8K, and HellaSwag. Green indicates improvement or minimal degradation; red indicates significant capability loss. Yi-1.5-9B with Heretic shows notable GSM8K degradation ($-26.5\%$), while DECCP and ErisForge maintain near-baseline performance.}
\label{fig:benchmark_heatmap}
\end{figure}

\subsection{Tool Comparison Summary}

Table~\ref{tab:toolsummary} presents mean capability changes by tool across the three models with complete baselines (DeepSeek-7B, Mistral-7B, Yi-1.5-9B).

\begin{table}[H]
\centering
\caption{Mean Capability Change by Tool (Percentage Points)}
\label{tab:toolsummary}
\begin{tabular}{lccc}
\toprule
\textbf{Tool} & \textbf{Avg MMLU $\Delta$ (pp)} & \textbf{Avg GSM8K $\Delta$ (pp)} & \textbf{Avg HellaSwag $\Delta$ (pp)} \\
\midrule
Heretic & $-0.78$ & $-7.81$ & $-0.56$ \\
DECCP & $-0.61$ & $-0.13$ & $-0.25$ \\
ErisForge & $-0.12$ & $-0.28$ & $-0.08$ \\
\bottomrule
\multicolumn{4}{p{0.92\textwidth}}{\footnotesize Mean of (Abliterated $-$ Base), averaged over DeepSeek-7B, Mistral-7B, Yi-1.5-9B.} \\
\end{tabular}
\end{table}

\begin{figure}[H]
\centering
\includegraphics[width=0.9\textwidth]{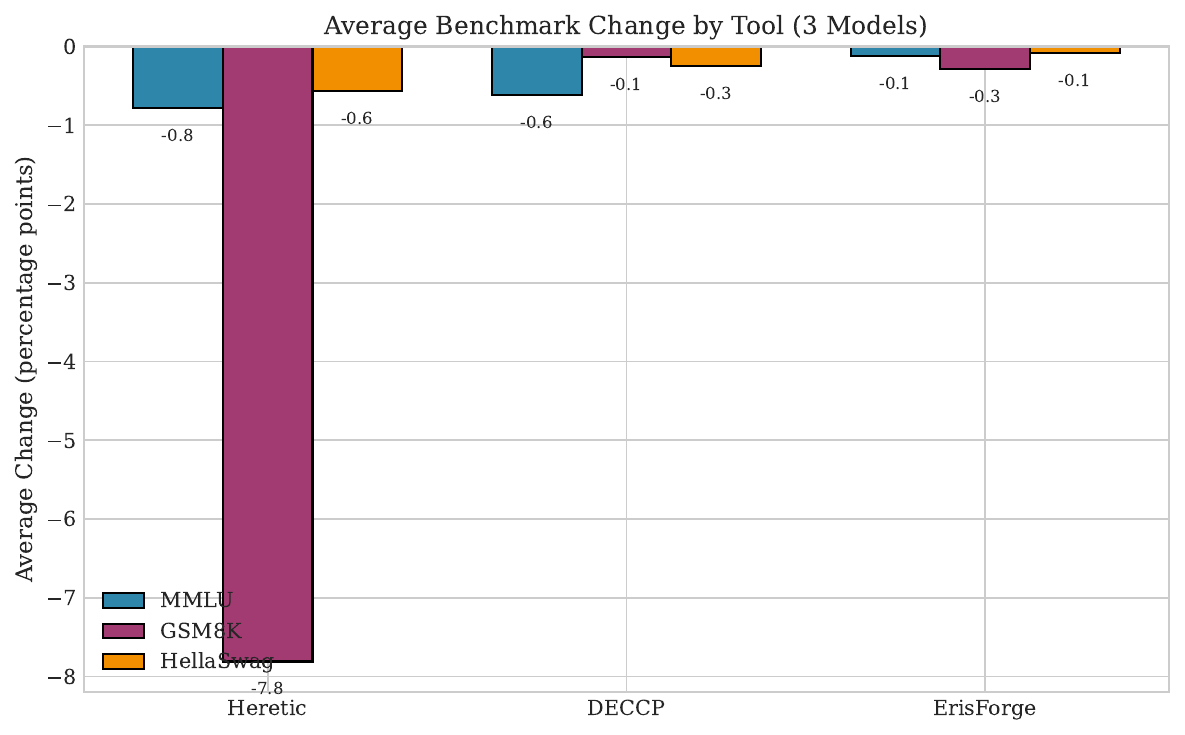}
\caption{Average benchmark change by tool across three models (DeepSeek-7B, Mistral-7B, Yi-1.5-9B). ErisForge demonstrated the best capability preservation across all benchmarks, followed by DECCP. Heretic showed significant GSM8K degradation on average ($-7.81$ pp) driven primarily by Yi-1.5-9B results.}
\label{fig:tool_comparison}
\end{figure}

\subsection{Model Coverage}

Table~\ref{tab:coverage} presents the complete model-tool coverage matrix across all sixteen tested models.

\begin{table}[H]
\centering
\caption{Model-Tool Coverage Matrix}
\label{tab:coverage}
\begin{tabular}{lccccc}
\toprule
\textbf{Model} & \textbf{Params} & \textbf{Heretic} & \textbf{DECCP} & \textbf{FailSpy} & \textbf{ErisForge} \\
\midrule
Llama-3.1-8B-Instruct & 8B & \checkmark & \checkmark & \checkmark & \checkmark \\
Mistral-7B-Instruct-v0.3 & 7B & \checkmark & \checkmark & \checkmark & \checkmark \\
Qwen2.5-7B-Instruct & 7B & \checkmark & \checkmark & \checkmark & \checkmark \\
Gemma-2-9B-it & 9B & \checkmark & \checkmark & \checkmark & \checkmark \\
Gemma-7B-it & 8B & \checkmark & \checkmark & \checkmark & \checkmark \\
StableLM-2-12B-chat & 12B & \checkmark & \checkmark & $\dagger$ & \checkmark \\
Yi-1.5-9B-Chat & 9B & \checkmark & \checkmark & $\dagger$ & \checkmark \\
Zephyr-7B-beta & 7B & \checkmark & \checkmark & $\dagger$ & \checkmark \\
deepseek-llm-7b-chat & 7B & \checkmark & \checkmark & $\dagger$ & \checkmark \\
OpenChat-3.5-0106 & 7B & \checkmark & \checkmark & $\dagger$ & $\times$ \\
Qwen3-8B & 8B & \checkmark & \checkmark & $\dagger$ & --- \\
Vicuna-7B-v1.5 & 7B & \checkmark & --- & $\dagger$ & $\times$ \\
InternLM2.5-7B-chat & 7B & \checkmark & --- & $\dagger$ & $\times$ \\
Falcon-Mamba-7B-instruct & 7B & \checkmark & $\ddagger$ & $\ddagger$ & $\ddagger$ \\
Phi-3-small-8k-instruct & 7B & \checkmark & --- & $\dagger$ & --- \\
Qwen3-14B & 14B & \checkmark & --- & $\dagger$ & --- \\
\bottomrule
\multicolumn{6}{l}{\footnotesize \checkmark~Successful \quad $\times$~Failed \quad ---~Not tested} \\
\multicolumn{6}{l}{\footnotesize $\dagger$~TransformerLens lacks model support \quad $\ddagger$~Architecturally incompatible (Mamba SSM)}
\end{tabular}
\end{table}

\begin{figure}[H]
\centering
\includegraphics[width=\textwidth]{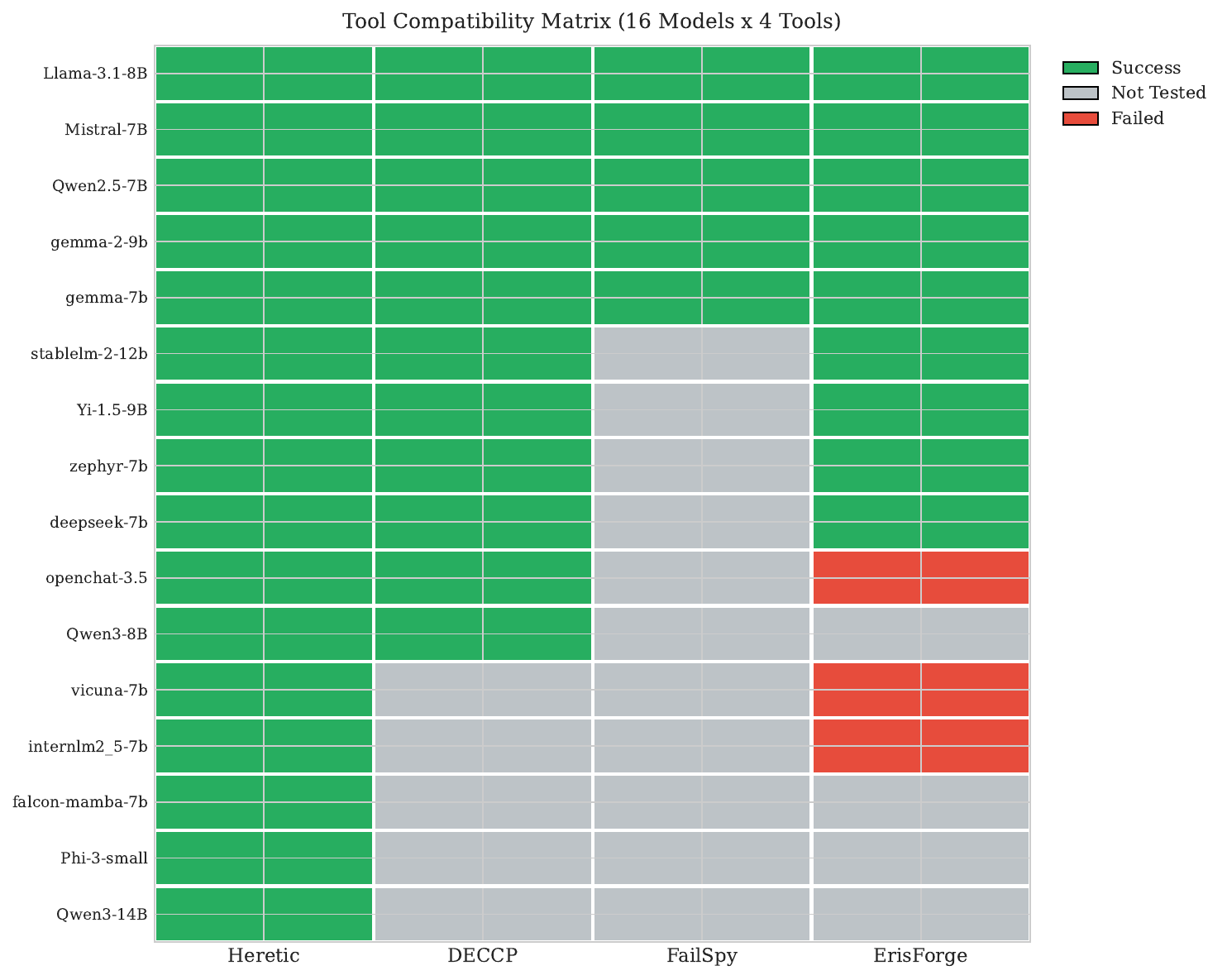}
\caption{Tool compatibility matrix across sixteen models. Heretic achieved universal compatibility (16/16 models), followed by DECCP (11/16). FailSpy's TransformerLens dependency limited compatibility to 5 models, while ErisForge succeeded on 9 models with some failures on models using non-standard architectures. Gray indicates not tested; Mamba SSM models are architecturally incompatible with transformer-only tools (see Table~\ref{tab:coverage}).}
\label{fig:coverage_matrix}
\end{figure}

\textbf{Coverage Totals}: Heretic: 16/16 (100\%), DECCP: 11/16 (69\%), FailSpy: 5/16 (31\%), ErisForge: 9/16 (56\%).

\section{Discussion}

\subsection{Tool Selection Considerations}

Based on this evaluation, the following patterns emerged across different use contexts:

\textbf{Quality-compliance optimization}: Heretic's Bayesian optimization produced the lowest observed KL divergence (minimum 0.043; range 0.043--1.646) but required the longest processing time ($\sim$45 min).

\textbf{Capability preservation priority}: ErisForge and DECCP showed the lowest benchmark degradation (avg GSM8K $\Delta$: $-0.28$ pp and $-0.13$ pp respectively).

\textbf{Mamba/hybrid architectures}: Only Heretic successfully processed SSM-containing models (falcon-mamba-7b-instruct).

\textbf{Processing efficiency}: DECCP completed in $\sim$2 min vs Heretic's $\sim$45 min (20$\times$ faster).

\subsection{Hypothesis Evaluation}

\textbf{H1 (Model Dependence)} is supported by substantial variance in both KL divergence and residual refusals across models (Table~\ref{tab:heretic}). Across these eight Heretic-evaluated models, KL divergence and remaining refusals are strongly correlated (Pearson $r=0.87$, $p=0.005$, $n=8$), consistent with a coupled trade-off between distribution shift and refusal suppression (Figure~\ref{fig:kl_vs_refusal}).

\textbf{H2 (Capability Preservation)} is supported on the benchmarked subset: single-pass methods (DECCP/ErisForge) show smaller average benchmark changes than Heretic (Table~\ref{tab:toolsummary}), and the largest divergence appears on GSM8K (Table~\ref{tab:benchmarks}).

\textbf{H3 (Alignment Sensitivity, Case Study)} is consistent with the DPO-only Zephyr-7B-beta exhibiting the lowest residual refusals and highest ASR under Heretic (Table~\ref{tab:heretic}). However, stronger conclusions require additional DPO-only models and consistent baseline coverage.

\subsection{Limitations}

Several limitations should inform interpretation of these findings:

\textbf{Statistical Rigor}: Due to computational constraints, the primary results reflect single experimental runs rather than multiple independent trials with variance estimation. While Heretic's Bayesian optimization internally evaluates multiple configurations (50 trials), each final abliterated model represents a single optimization outcome. This design choice was driven by practical considerations: full benchmark evaluation of a single 7B model requires approximately 4 to 6 GPU-hours, making triplicate runs across 16 models and 4 tools prohibitively expensive ($>$1,500 GPU-hours). To mitigate this limitation, deterministic execution was ensured (fixed seeds, deterministic CUDA where possible) and complete reproducibility artifacts are provided enabling independent replication. Future work with dedicated compute resources should incorporate proper bootstrap confidence intervals and statistical significance testing across multiple runs. Notably, the observed effect sizes for tool comparisons (e.g., Yi-1.5-9B GSM8K: Heretic 52.08\% vs DECCP 72.40\%, $\Delta = 20.32$ pp) substantially exceed benchmark uncertainty (GSM8K 95\% CI half-width $\sim$2.4--2.7 pp; Appendix), suggesting robust underlying differences despite single-run methodology.

\textbf{Metric Coverage}: The capability evaluation focuses on three benchmarks (MMLU, GSM8K, HellaSwag). Due to tool compatibility constraints and computational budget, baseline-relative cross-tool comparisons were limited to three models (DeepSeek-7B, Mistral-7B, Yi-1.5-9B); Zephyr-7B-beta is included without a base-model baseline. While compatibility was assessed across 16 models, quantitative metrics (KL divergence, refusals, and benchmark deltas) are not available for every model-tool pair. Additionally, these benchmarks may not capture all forms of capability degradation (e.g., instruction-following quality, long-context coherence, or domain-specific expertise).

\textbf{Refusal/ASR Validity}: Refusal rate and ASR rely on a marker-based detection heuristic and may misclassify outputs. False positives can occur when responses mention safety language (e.g., ``illegal'') while still providing actionable content; false negatives can occur when refusals are phrased without the selected markers. Additionally, refusal rates are estimated from $n=100$ prompts per model and therefore have non-trivial sampling uncertainty (Appendix). As a lightweight validity check, we cross-validated the marker heuristic using an independent RoBERTa-based refusal/disclaimer classifier trained on Chatbot Arena responses \cite{llmrefusalclassifier}. Disagreements were dominated by disclaimer-style outputs, suggesting that marker-based ASR can be conservative (i.e., under-estimate refusal bypass) when models retain safety language while still answering (Appendix Table~\ref{tab:heuristic_autoval}). As a result, ASR should be interpreted as a proxy for refusal suppression rather than a definitive measure of harmful content generation.

\textbf{Tool Configuration and Budget}: Tools were evaluated using their recommended default settings. This can conflate algorithmic differences with configuration choices and compute budgets (e.g., multi-trial optimization vs single-pass ablation). Future work should compare tools under matched compute budgets or matched refusal-rate targets (Pareto curves of refusal suppression, KL divergence, and capability).

\textbf{Model Selection}: The sixteen-model evaluation, while architecturally diverse, represents a subset of available instruction-tuned models. Notably absent are larger models ($>$14B parameters), mixture-of-experts architectures (e.g., Mixtral), and models from some providers (Claude, GPT). Results may not generalize to these architectures, particularly for models employing novel safety training techniques or different internal representational structures.

\textbf{Long-term and Deployment Effects}: This study evaluates immediate post-abliteration performance under controlled benchmark conditions. Real-world deployment introduces variables not captured here: multi-turn conversation drift, distribution shift from user prompts differing from evaluation sets, and potential emergent behaviors under extended use. Whether abliterated models maintain stability over thousands of interactions remains uncharacterized.

\section{Conclusion}

This study presented a systematic comparison of four LLM abliteration tools (Heretic, DECCP, ErisForge, and FailSpy/abliterator) across sixteen instruction-tuned models, reporting tool compatibility across all models and capability preservation benchmarks on a subset due to tool and compute constraints. The evaluation yielded the following findings:

\textbf{1. Capability preservation varied significantly by tool.} In the benchmarked subset, ErisForge exhibited the lowest average benchmark degradation (MMLU: $-0.12$ pp, GSM8K: $-0.28$ pp, HellaSwag: $-0.08$ pp), followed by DECCP (GSM8K: $-0.13$ pp). Heretic demonstrated model-specific variance, with minimal impact on Mistral-7B ($-0.12$ pp avg) but substantial GSM8K degradation on Yi-1.5-9B ($-18.81$ pp; $-26.5\%$ relative).

\textbf{2. Mathematical reasoning showed highest sensitivity to abliteration.} GSM8K scores exhibited the largest variance across tools and models (range: $+1.51$ pp to $-18.81$ pp), suggesting potential overlap between mathematical reasoning circuits and refusal-related representations.

\textbf{3. Automated optimization produced quality-speed tradeoffs.} Heretic's Bayesian optimization achieved a minimum KL divergence of 0.043 (range: 0.043--1.646) with explicit refusal tracking, but required $\sim$45 minutes per model. DECCP and ErisForge completed processing in $\sim$2--20 minutes with more predictable capability preservation.

\textbf{4. Model compatibility differed substantially across tools.} Heretic processed all 16 tested models (100\% coverage), while DECCP supported 11/16 (69\%), ErisForge 9/16 (56\%), and FailSpy 5/16 (31\%).

\textbf{5. Model resilience varied across architectures.} Mistral-7B exhibited consistent stability across all tools (avg $\Delta$: $-0.12$ to $-0.61$ pp), while Yi-1.5-9B showed high sensitivity to Heretic's kernel-weighted approach but improved GSM8K performance ($+1.51$ pp) with DECCP.

These findings establish empirical baselines for tool comparison and provide reproducible protocols for evaluating abliteration approaches.

\textbf{Future Directions.} Several promising extensions merit investigation. Recent contributions to the llm-abliteration repository \cite{jimplus} have extended support to Mixture-of-Experts (MoE) architectures, enabling abliteration of models such as Mixtral and DeepSeek-MoE that were previously incompatible with transformer-focused tools. Additionally, the introduction of an ``invert'' option enables directional \emph{addition} rather than removal, potentially amplifying specific behavioral traits---a capability that opens novel applications in steering model personality, enhancing helpfulness, or studying the symmetric effects of representation engineering. Systematic evaluation of these extensions across MoE architectures and bidirectional modification scenarios represents a natural continuation of this work.

\section*{Acknowledgments}

The author extends heartfelt gratitude to Dr.\ Michael Lee at the University of Nevada, Las Vegas for inspiring this line of research into LLM safety. His mentorship opened the door to a fascinating intersection of security, computational neuroscience, and large language model understanding that has reignited a passion for this field. Thank you, Mike. The author also acknowledges computational resources provided by DeepNeuro.ai for research infrastructure, NVIDIA for GPU resources and tooling, and the open-source community for the abliteration tools evaluated in this study.

\section*{Data and Code Availability}

Benchmark results, experimental configurations, and abliteration directions are available at \url{https://github.com/ricyoung/abliteration-comparison}. Select abliterated model weights for permissively-licensed base models are available on Hugging Face at \url{https://huggingface.co/collections/richardyoung/uncensored-and-abliterated-llms-69267b2d66ae2cc21c4b0e99}. Abliterated weights for models with restrictive licenses are available upon request to verified researchers for reproducibility purposes.

\section*{Competing Interests}

The author declares no competing interests.

\section*{Appendix: Additional Uncertainty and Validity Checks}

\subsection*{A. Benchmark 95\% Confidence Intervals}

\begin{table}[H]
\centering
\caption{Benchmark Scores (\% $\pm$ 95\% Confidence Interval)}
\label{tab:bench_ci}
\scriptsize
\begin{tabular}{@{}llccc@{}}
\toprule
\textbf{Model} & \textbf{Variant} & \textbf{MMLU} & \textbf{GSM8K} & \textbf{HellaSwag} \\
\midrule
DeepSeek-7B & Base & 49.44\% $\pm$ 0.79 & 44.58\% $\pm$ 2.68 & 77.84\% $\pm$ 0.81 \\
 & Heretic & 48.95\% $\pm$ 0.79 & 40.11\% $\pm$ 2.65 & 77.62\% $\pm$ 0.82 \\
 & DECCP & 49.05\% $\pm$ 0.79 & 43.59\% $\pm$ 2.68 & 77.99\% $\pm$ 0.81 \\
 & ErisForge & 49.43\% $\pm$ 0.79 & 44.35\% $\pm$ 2.68 & 77.69\% $\pm$ 0.81 \\
\midrule
Mistral-7B & Base & 59.74\% $\pm$ 0.76 & 48.52\% $\pm$ 2.70 & 83.28\% $\pm$ 0.73 \\
 & Heretic & 59.46\% $\pm$ 0.76 & 48.37\% $\pm$ 2.70 & 83.36\% $\pm$ 0.73 \\
 & DECCP & 58.98\% $\pm$ 0.76 & 47.61\% $\pm$ 2.70 & 83.12\% $\pm$ 0.73 \\
 & ErisForge & 59.42\% $\pm$ 0.76 & 48.29\% $\pm$ 2.70 & 83.35\% $\pm$ 0.73 \\
\midrule
Yi-1.5-9B & Base & 68.02\% $\pm$ 0.74 & 70.89\% $\pm$ 2.45 & 78.62\% $\pm$ 0.80 \\
 & Heretic & 66.46\% $\pm$ 0.74 & 52.08\% $\pm$ 2.70 & 77.08\% $\pm$ 0.82 \\
 & DECCP & 67.33\% $\pm$ 0.74 & 72.40\% $\pm$ 2.41 & 77.87\% $\pm$ 0.81 \\
 & ErisForge & 67.99\% $\pm$ 0.74 & 70.51\% $\pm$ 2.46 & 78.46\% $\pm$ 0.80 \\
\midrule
Zephyr-7B-beta & Heretic & 58.50\% $\pm$ 0.77 & 33.36\% $\pm$ 2.55 & 82.90\% $\pm$ 0.74 \\
 & DECCP & 58.28\% $\pm$ 0.77 & 33.21\% $\pm$ 2.54 & 82.05\% $\pm$ 0.75 \\
\bottomrule
\multicolumn{5}{p{0.92\textwidth}}{\footnotesize Each cell reports score $\pm 95\%$ CI, computed as $\pm 1.96 \times \text{stderr}$ using standard errors from lm-evaluation-harness (GSM8K uses strict-match exact match).} \\
\end{tabular}
\end{table}

\subsection*{B. Refusal/ASR 95\% Confidence Intervals}

\begin{table}[H]
\centering
\caption{Heretic Refusal Rates with 95\% Wilson Confidence Intervals ($n=100$ harmful prompts)}
\label{tab:refusal_ci}
\footnotesize
\begin{tabular}{lccc}
\toprule
\textbf{Model} & \textbf{Refusals} & \textbf{ASR} & \textbf{ASR 95\% CI (Wilson)} \\
\midrule
Zephyr-7B-beta & 2/100 & 98\% & 93.0--99.4 \\
DeepSeek-7B-chat & 16/100 & 84\% & 75.6--89.9 \\
Mistral-7B-v0.3 & 16/100 & 84\% & 75.6--89.9 \\
Llama-3.1-8B & 24/100 & 76\% & 66.8--83.3 \\
Qwen3-8B & 25/100 & 75\% & 65.7--82.5 \\
Yi-1.5-9B & 25/100 & 75\% & 65.7--82.5 \\
Qwen2.5-7B & 42/100 & 58\% & 48.2--67.2 \\
StableLM-2-12B & 54/100 & 46\% & 36.6--55.7 \\
\bottomrule
\multicolumn{4}{p{0.92\textwidth}}{\footnotesize Confidence intervals reflect prompt-sampling uncertainty only and do not account for potential heuristic misclassification (see Limitations).} \\
\end{tabular}
\end{table}

\subsection*{C. Automated Refusal Heuristic Cross-Validation}

To bound refusal/ASR heuristic misclassification without human labeling, we compared the marker-based refusal heuristic used in this paper against an independent RoBERTa-based refusal/disclaimer classifier \cite{llmrefusalclassifier} on existing generations from a separate harmful-prompt set (150 prompts per model, six abliterated models; $n=900$ responses). The classifier distinguishes refusals from disclaimers; for this comparison, disclaimers are treated as non-refusals (i.e., counted toward ASR) because they attempt to answer while citing constraints. Table~\ref{tab:heuristic_autoval} reports agreement metrics for the full marker set (paper default) and a stricter variant that drops generic safety-topic substrings (e.g., ``illegal'', ``unethical'') to illustrate sensitivity.

\begin{table}[H]
\centering
\caption{Marker Heuristic vs Refusal/Disclaimer Classifier (Aggregate, $n=900$ responses)}
\label{tab:heuristic_autoval}
\scriptsize
\begin{tabular}{lcccc}
\toprule
\textbf{Marker Set} & \textbf{ASR (Markers)} & \textbf{ASR (Classifier)} & \textbf{Prec. (Refusal)} & \textbf{Recall (Refusal)} \\
\midrule
Full (paper default) & 72.2\% (69.2--75.0) & 95.7\% (94.1--96.8) & 11.2\% & 71.8\% \\
Strict (drop generic safety words) & 96.8\% (95.4--97.7) & 95.7\% (94.1--96.8) & 37.9\% & 28.2\% \\
\bottomrule
\multicolumn{5}{p{0.92\textwidth}}{\footnotesize ASR values report non-refusal rate with Wilson 95\% confidence intervals; precision/recall treat the classifier refusal label as reference. This is a cross-validation signal rather than ground truth: both methods can misclassify, and neither directly measures actionable harmful compliance.} \\
\end{tabular}
\end{table}

\begin{table}[H]
\centering
\caption{Per-Model ASR Discrepancy (Full Marker Set, $n=150$ responses per model)}
\label{tab:heuristic_autoval_models}
\scriptsize
\begin{tabular}{lcccc}
\toprule
\textbf{Model} & \textbf{ASR (Markers)} & \textbf{ASR (Classifier)} & \textbf{$\Delta$ (pp)} & \textbf{Disclaimers (Classifier)} \\
\midrule
DeepSeek-7B-chat & 55.3\% & 84.7\% & +29.3 & 37.3\% \\
Mistral-7B-v0.3 & 76.7\% & 99.3\% & +22.7 & 15.3\% \\
Yi-1.5-9B & 79.3\% & 99.3\% & +20.0 & 16.0\% \\
Zephyr-7B-beta & 85.3\% & 97.3\% & +12.0 & 12.0\% \\
Llama-3.1-8B & 86.7\% & 93.3\% & +6.7 & 22.7\% \\
Qwen2.5-7B & 50.0\% & 100.0\% & +50.0 & 22.7\% \\
\bottomrule
\multicolumn{5}{p{0.92\textwidth}}{\footnotesize $\Delta = \text{ASR}_{\text{classifier}} - \text{ASR}_{\text{markers}}$; positive values indicate the marker heuristic is more conservative (flags more outputs as refusals). ``Disclaimers'' are classifier labels indicating safety/capability caveats while still attempting to answer (counted as non-refusal for ASR in this table).} \\
\end{tabular}
\end{table}



\begin{thebibliography}{99}

\bibitem{arditi2024refusal}
Arditi, A., Obeso, O., Shlegeris, B., et al. ``Refusal in Language Models Is Mediated by a Single Direction.'' arXiv:2406.11717 (2024).

\bibitem{mikolov2013distributed}
Mikolov, T., Sutskever, I., Chen, K., Corrado, G. S., \& Dean, J. ``Distributed representations of words and phrases and their compositionality.'' NeurIPS (2013).

\bibitem{heretic}
p-e-w. ``Heretic: Fully automatic censorship removal for language models.'' GitHub. \url{https://github.com/p-e-w/heretic}

\bibitem{failspy}
FailSpy. ``abliterator: Ablate features in transformer-based LLMs.'' GitHub. \url{https://github.com/FailSpy/abliterator}

\bibitem{jimplus}
Lai, J. ``llm-abliteration: Fast abliteration without TransformerLens.'' GitHub. \url{https://github.com/jim-plus/llm-abliteration}

\bibitem{llmrefusalclassifier}
Human-CentricAI. ``LLM-Refusal-Classifier: RoBERTa-based refusal/disclaimer classifier.'' Hugging Face Hub (2025). \url{https://huggingface.co/Human-CentricAI/LLM-Refusal-Classifier}

\bibitem{erisforge}
Tsadoq. ``ErisForge.'' GitHub. \url{https://github.com/Tsadoq/ErisForge}

\bibitem{zou2023representation}
Zou, A., et al. ``Representation engineering: A top-down approach to AI transparency.'' arXiv:2310.01405 (2023).

\bibitem{turner2023activation}
Turner, A. M., et al. ``Steering Language Models With Activation Engineering.'' arXiv:2308.10248 (2023).

\bibitem{labonne2024abliteration}
Labonne, M. ``Uncensor any LLM with abliteration.'' HuggingFace Blog (2024).

\bibitem{lai2024projected}
Lai, J. ``Projected Abliteration: Norm-preserving biprojected abliteration.'' HuggingFace Blog (2024). \url{https://huggingface.co/blog/grimjim/projected-abliteration}

\bibitem{rafailov2023dpo}
Rafailov, R., et al. ``Direct preference optimization: Your language model is secretly a reward model.'' NeurIPS (2023).


\bibitem{abushairah2025defense}
Abu Shairah, M., et al. ``An Embarrassingly Simple Defense Against LLM Abliteration Attacks.'' arXiv:2505.19056 (2025).

\bibitem{agnihotri2025safety}
Agnihotri, S., Jakubassa, J., Dey, P., Goyal, S., Schiele, B., Radhakrishnan, V. B., \& Keuper, M. ``A Granular Study of Safety Pretraining under Model Abliteration.'' arXiv:2510.02768 (2025).

\bibitem{piras2025som}
Piras, G., Mura, R., Brau, F., Oneto, L., Roli, F., \& Biggio, B. ``SOM Directions are Better than One: Multi-Directional Refusal Suppression in Language Models.'' arXiv:2511.08379 (2025).

\bibitem{wang2025refusal}
Wang, X., Wang, M., Liu, Y., Sch\"utze, H., \& Plank, B. ``Refusal Direction is Universal Across Safety-Aligned Languages.'' arXiv:2505.17306 (2025).

\bibitem{wollschlaeger2025geometry}
Wollschl\"ager, T., Elstner, J., Geisler, S., Cohen-Addad, V., G\"unnemann, S., \& Gasteiger, J. ``The Geometry of Refusal in Large Language Models: Concept Cones and Representational Independence.'' arXiv:2502.17420 (2025).

\bibitem{zhao2025harmfulness}
Zhao, J., Huang, J., Wu, Z., Bau, D., \& Shi, W. ``LLMs Encode Harmfulness and Refusal Separately.'' arXiv:2507.11878 (2025).


\bibitem{zou2023universal}
Zou, A., et al. ``Universal and transferable adversarial attacks on aligned language models.'' arXiv:2307.15043 (2023).

\bibitem{wei2023jailbroken}
Wei, A., Haghtalab, N., \& Steinhardt, J. ``Jailbroken: How does LLM safety training fail?'' arXiv:2307.02483 (2023).

\bibitem{chao2024jailbreakbench}
Chao, P., et al. ``JailbreakBench: An Open Robustness Benchmark for Jailbreaking Large Language Models.'' arXiv:2404.01318 (2024).

\bibitem{mazeika2024harmbench}
Mazeika, M., et al. ``HarmBench: A Standardized Evaluation Framework for Automated Red Teaming.'' arXiv:2402.04249 (2024).


\bibitem{wang2024alignment}
Wang, J., et al. ``A Comprehensive Survey of LLM Alignment Techniques: RLHF, RLAIF, PPO, DPO and More.'' arXiv:2407.16216 (2024).

\bibitem{ouyang2022instructgpt}
Ouyang, L., et al. ``Training language models to follow instructions with human feedback.'' NeurIPS (2022).

\bibitem{bai2022constitutional}
Bai, Y., et al. ``Constitutional AI: Harmlessness from AI feedback.'' arXiv:2212.08073 (2022).

\bibitem{qi2024safety}
Qi, X., et al. ``Safety alignment should be made more than just a few tokens deep.'' arXiv:2406.05946 (2024).

\bibitem{inan2023llamaguard}
Inan, H., et al. ``Llama Guard: LLM-based Input-Output Safeguard for Human-AI Conversations.'' arXiv:2312.06674 (2023).


\bibitem{tunstall2023zephyr}
Tunstall, L., et al. ``Zephyr: Direct Distillation of LM Alignment.'' arXiv:2310.16944 (2023).

\bibitem{cyberey2025steering}
Cyberey, H. \& Evans, D. ``Steering the CensorShip: Uncovering Representation Vectors for LLM Thought Control.'' arXiv:2504.17130 (2025).

\bibitem{jain2024mechanics}
Jain, S., et al. ``What makes and breaks safety fine-tuning? A mechanistic study.'' NeurIPS (2024).


\bibitem{zou2024circuitbreaker}
Zou, A., et al. ``Improving alignment robustness with circuit breakers.'' arXiv:2406.04313 (2024).

\bibitem{park2024linear}
Park, K., et al. ``The Linear Representation Hypothesis and the Geometry of Large Language Models.'' arXiv:2311.03658 (2023).


\bibitem{hendrycks2021mmlu}
Hendrycks, D., et al. ``Measuring Massive Multitask Language Understanding.'' ICLR (2021).

\bibitem{cobbe2021gsm8k}
Cobbe, K., et al. ``Training Verifiers to Solve Math Word Problems.'' arXiv:2110.14168 (2021).

\bibitem{zellers2019hellaswag}
Zellers, R., et al. ``HellaSwag: Can a Machine Really Finish Your Sentence?'' ACL (2019).

\end{thebibliography}
\end{document}